\documentclass[10pt,twocolumn,letterpaper]{article}

\usepackage{wacv}
\makeatletter
\@namedef{ver@everyshi.sty}{}
\makeatother
\usepackage{tikz}
\usepackage{times}
\usepackage{epsfig}
\usepackage{graphicx}
\usepackage{amsmath}
\usepackage{amssymb}
\usepackage{booktabs}
\usepackage[accsupp]{axessibility}

\usepackage{bm}

\usepackage{pifont}
\newcommand{\cmark}{\ding{51}}%
\newcommand{\xmark}{\ding{55}}%
\usepackage{multirow}
\usepackage[normalem]{ulem}
\useunder{\uline}{\ul}{}

\usepackage{pgfplots} 
\usepackage{subcaption}
\usepackage{xcolor}

\definecolor{dark_blue}{RGB}{  0, 45, 106}
\definecolor{light_blue}{RGB}{ 220, 238, 243}

\def\wacvPaperID{1526} 

\wacvalgorithmstrack   

\wacvfinalcopy 

\ifwacvfinal
\usepackage[breaklinks=true,bookmarks=false]{hyperref}
\else
\usepackage[pagebackref=true,breaklinks=true,colorlinks,bookmarks=false]{hyperref}
\fi

\makeatletter
\def\thanks#1{\protected@xdef\@thanks{\@thanks
        \protect\footnotetext{#1}}}
\makeatother

\begin{document}

\title{Intention-Conditioned Long-Term Human Egocentric Action Anticipation}


\author{Esteve Valls Mascar\'o $^{1}$ \and Hyemin Ahn $^{2}$ \and Dongheui Lee $^{1,3}$
\thanks{$^{1}$Esteve Valls Mascaro and Dongheui Lee are with Autonomous Systems, Technische Universität Wien (TU Wien), Vienna, Austria (e-mail: \texttt{\{esteve.valls.mascaro, dongheui.lee\}@tuwien.ac.at}).}%
\thanks{$^{2}$Hyemin Ahn is with Artificial Intelligence Graduate School (AIGS), Ulsan National Institute of Science and Technology (UNIST), Ulsan, Korea (e-mail: \texttt{hyemin.ahn@unist.ac.kr}).}%
\thanks{$^{3}$Dongheui Lee is also with the Institute of Robotics and Mechatronics, German Aerospace Center, Wessling, Germany.}%
\thanks{This work is funded by Marie Sklodowska-Curie Action Horizon 2020 (Grant agreement No. 955778) for project 'Personalized Robotics as Service Oriented Applications' (PERSEO).}
}

\maketitle
\thispagestyle{empty}

\begin{abstract}
To anticipate how a person would act in the future, it is essential to understand the human intention since it guides the subject towards a certain action. In this paper, we propose a hierarchical architecture which assumes a sequence of human action (low-level) can be driven from the human intention (high-level). Based on this, we deal with long-term action anticipation task in egocentric videos. Our framework first extracts this low- and high-level human information over the observed human actions in a video through a Hierarchical Multi-task Multi-Layer Perceptrons Mixer (H3M). Then, we constrain the uncertainty of the future through an Intention-Conditioned Variational Auto-Encoder (I-CVAE) that generates multiple stable predictions of the next actions that the observed human might perform. By leveraging human intention as high-level information, we claim that our model is able to anticipate more time-consistent actions in the long-term, thus improving the results over the baseline in Ego4D dataset. This work results in the state-of-the-art for Long-Term Anticipation (LTA) task in Ego4D by providing more plausible anticipated sequences, improving the anticipation scores of nouns and actions. Our work ranked first in both CVPR@2022 and ECCV@2022 Ego4D LTA Challenge.

\end{abstract}

\section{Introduction}
In our everyday life, reasoning about next actions is essential before executing a certain complex task. Humans can project oneself into the future through a constructive imagination system, that allows them to anticipate the future actions of themselves as well as others. Based on this, various applications such as task and motion planning (TAMP), or human-robot-collaboration can be realized. For instance, after understanding that someone is about to make a salad, we can first anticipate his/her next actions to later assist this person to prepare ingredients. This field of research in computer vision, known as Long-Term Anticipation (LTA) of human actions, aims at forecasting the actions that a human will perform most likely based on the past observations.

The fundamental challenge of LTA of human actions is the inherent uncertainty of the future. The uniqueness of the human being results in high variability of how each of us executes a certain task. Moreover, this behaviour may vary for the same individual at different moments and also depending on the environment. However, despite the theoretical high variability of possible predictions, the future often has only a limited number of plausible outcomes. Inspired by this, we hypothesize that the arbitrariness of future events can be narrowed down through conditioning on past observations, which would imply the context of the whole task. For instance, if we observe a human cutting a tomato and we know that the human's intention is to make a salad, the variety of ingredients that the human might interact with can be reduced. This human intention, defined as a high-level hypothesis, conditions the human behavior and reduces the variability and uncertainty of the future.

\begin{figure*}
\centering
\includegraphics[width=0.85\textwidth]{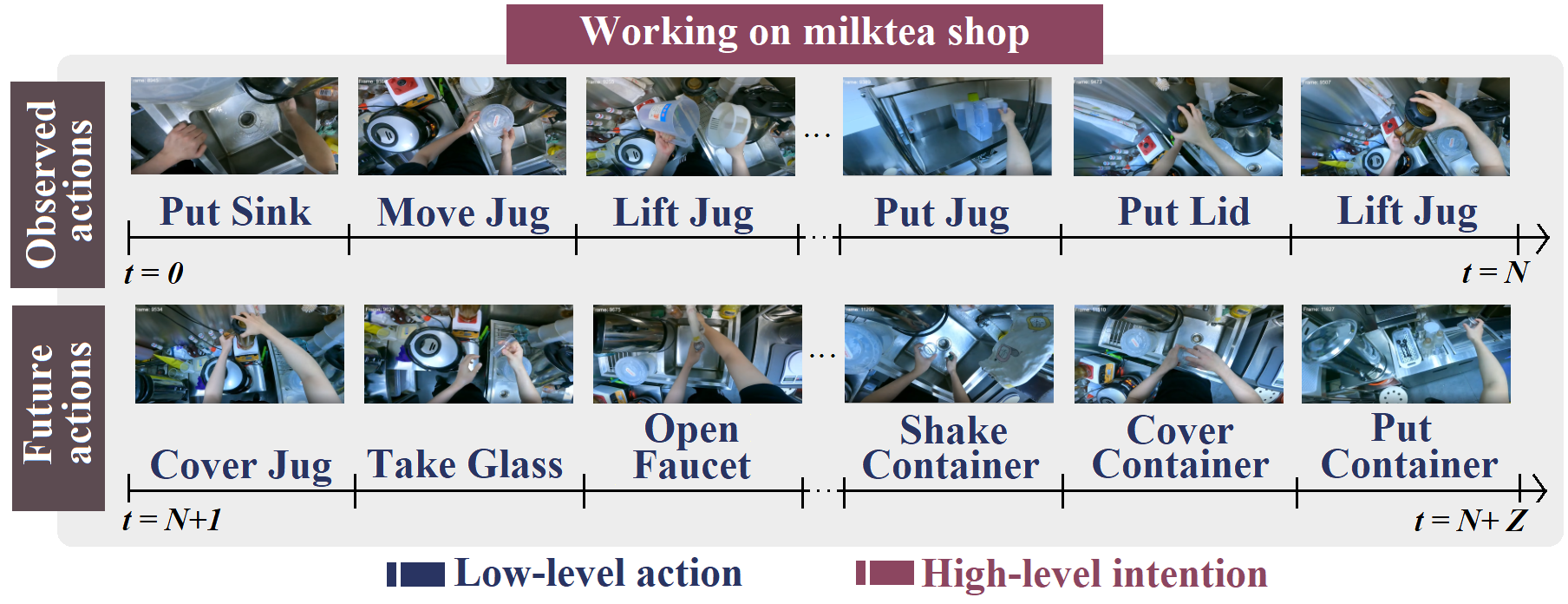}
\caption{\textbf{An example of the hierarchy structure of a human task}. Egocentric sequence of videos of a human `Working on milktea shop' (in purple, describing the high-level human intention) from Ego4d \cite{ego4d}. In blue, a sequence of low-level actions labels performed by the camera wearer is shown. This paper proposes a methodology that understands a human task based on this hierarchical structure. Our model extracts a high-level human intention information and $N$ action labels from the observed sequence of $N$ clips (first row) to facilitate the anticipation of low-level $Z$ actions in its future (second row).}
\label{fig:dataset_hierarchy}
\end{figure*}

Therefore, we develop a methodology that aims to constrain the variability of future actions based on the human intention estimated from past observations. We predict a hierarchical structure from a sequence of videos, each depicting a particular human action. From this given video clip sequence, we define two different levels of abstraction, as visualized in Fig. \ref{fig:dataset_hierarchy}. First, we explore the human intention as the highest concept that defines the ultimate objective pursued by the human while performing a task. Second, we make use of the sequence of low-level actions to perform a certain task. We aim to mimic a human's reasoning process when predicting which actions to perform given a high-level instruction: (i) determine the current status, or context, of the task by observing the steps already completed (past low-level actions), (ii) plan the next steps based on the ultimate goal (intention) of the task. Therefore, based on the historical sequence of low-level actions performed by a human, we propose to exploit the high-level human intention as a guidance that conditions the next actions that will be executed in the future.

Following the division of our methodology, we define our framework as a two-step. First, we propose a Hierarchical Multitask Multi-Layer Perceptrons (MLP) Mixer (H3M), to classify each observed video to an action label, as well as to extract the overall intention of the human. The MLP Mixer-based architecture \cite{mlpmixer} has been empirically shown to be an optimal model that uses the repeated MLP layers through temporal and spatial channels. Our H3M is designed as a multitask network \cite{multitask} to exploit dependencies between low-level actions and high-level intentions while making the network more efficient. Secondly, we design an Intention-Conditioned Variational Autoencoder (I-CVAE), to anticipate the user's future actions conditioned by the human intention and the observed past actions. Variational Autoencoder (VAE) \cite{vae} has been shown in \cite{generative_lta_vae_2, generative_lta_vae_1} to effectively model the human action sequence distribution. However, to narrow the uncertainty of the future, our model is based on Conditional VAE (CVAE), inspired by \cite{actor}, leveraging the inferred human intention as a latent condition that provides a guideline to the model to anticipate the upcoming action sequence.

To demonstrate the effectiveness of our approach, we make use of the most diverse dataset of human videos currently available, Ego4D \cite{ego4d}, and in particular we evaluate our results in the LTA benchmark. Ego4D provides first-person videos of humans experiencing everyday activities around the world. In the case of the LTA task, it proposes to predict the future sequence of actions of the camera user from the untrimmed video of its past. Low-level actions have been annotated in videos, which are classified depending on a \texttt{scenario}, that we understand as \texttt{human intention}, as can be seen in Fig. \ref{fig:dataset_hierarchy}

Finally, we report quantitative results that lead our approach to win both CVPR@2022 and ECCV@2022 Ego4D Long-Term Action Anticipation (LTA) challenges. We extend the results with a detailed discussion based on our ablation study. To sum up, the contributions can be summarized as follows:
\begin{enumerate}
   \item It aims to extract the sequence of low-level actions as well as the human intention from a video through a multi-task hierarchical structure.
   \item It promotes the use of high-level intention as a condition to anticipate the future sequence of actions.
   \item It provides detailed analysis of long-term human action anticipation task, based on ablation studies, that aims to point out new research questions.
\end{enumerate}

\section{Related work}
Long-Term Anticipation (LTA) has been a fundamental challenge in the computer vision research community. In the following section, we discuss the most relevant research in that field. Then, we review several works for hierarchical extraction and generative models.

\subsection{Long-Term Anticipation}

Predicting human's future events has been covered in computer vision based on different tasks, such as generating skeleton motion \cite{actor, actformer, mdm}, predicting future human trajectory \cite{location_third_person, location2, location1} or generating a sequence of future action labels \cite{generative_lta_vae_2, generative_lta_vae_1, long_range_lta}. Due to the broadness of the literature, we will focus on works related to the action sequence anticipation task.

Existing literature \cite{generative_lta_randomsample, twostep_deterministic_lta, generative_lta_vae_2, generative_lta_vae_1, onestep_deterministic_lta, lta_latent_goal, long_range_lta} can be classified into two categories in terms of how to deal with the future. On the one hand, researchers aimed at simplifying the problem by modelling future as a deterministic function based on the observed video, regardless of future uncertainty \cite{twostep_deterministic_lta, onestep_deterministic_lta}. To overcome this simplification, two-step approaches were also taken into account in \cite{twostep_deterministic_lta}: Recurrent neural network (RNN) \cite{rnn} was used to first infer the action label sequence from the observed video, and then feed these classified actions into another RNN to forecast the future action sequence. However, \cite{onestep_deterministic_lta} argued that anticipating unseen sequence of future action labels directly from observed videos provides more contextual cues. 

On the other hand, inspired by the probabilistic nature of the future, \cite{generative_lta_randomsample} attempted to model the uncertainty through recursively sampling a future action from a softmax distribution obtained through a deterministic RNN based on previously predicted actions. To adapt the probabilistic essence in the architecture, Variational Autoencoder (VAE) \cite{vae} model, combined with point process models, was proposed in \cite{generative_lta_vae_1} to learn a latent distribution conditioned on the observed action sequence. Based on that, future actions are generated. \cite{generative_lta_vae_2} reused a VAE-approach but modelled a Multi-Head attention-based (MHA) Variational RNN to encode the latent distribution, which is concatenated in the RNN's hidden state to decode the next actions recursively.

Our work is also inspired by both generative approaches \cite{generative_lta_vae_2, generative_lta_vae_1} but attempts at minimizing the future uncertainty by adding human intention as a condition learned from the latent distribution. While experimental results \cite{long_range_lta} demonstrated the key role of recent actions for immediate future predictions, we claim that knowing the intention guides the intelligent agent through a better long-term anticipation.

More recent work such as \cite{lta_latent_goal} demonstrated the use of latent goals, obtained from the observed actions, as a feature representation used to anticipate the next action. However, this latent representation is not explainable as it does not consist of language-based labels. Moreover, \cite{lta_latent_goal} only attempted to select one next action based on RNN proposed candidates, then dealing with short-term anticipation. On the contrary, we propose a single high-level explainable label, the human intention, as a guidance for the anticipation of long-term sequence of low-level actions. This attempt to hierarchically model the future is inspired by \cite{hyperbolic_future}, which coped with the uncertainty by abstracting the level of prediction when the model confidence score was low. Our goal differs in nature, as we aim to always predict low-level actions, but encourage confidence to the model based on the high-level intention. 

To the best of our knowledge, our work is the first attempt to decompose the future in a two-level explainable hierarchical structure. This design allows to deal with time-uncertainty by top-down approaches: high-level intention is used for robust anticipation of the low-level actions.

\subsection{Generative Models}

Understanding and modelling the data distribution and generalizing it to unseen scenarios is key for a deep generative model. When generative models succeed in synthesizing plausible data, it is assumed that the model has learned the data distribution properly. Deep generative models can be applied to a wide range of domains, from the generation of images \cite{biggan, dalle2, imagen} to text \cite{gpt3, vae_text}, skeleton movement \cite{actor, actformer, mdm} and the synthesis of forthcoming actions \cite{generative_lta_vae_2, generative_lta_vae_1}.

Despite the enormous variety of works regarding generative models, a recent convention is to classify deep generative models into three different directions \cite{original_gan, dm, vae}. First, Generative Adversarial Networks (GANs) \cite{biggan, original_gan, actformer} benefit from adversarial training of two networks that contest to maximize its own objective function in opposed tasks, thus encouraging the opponent network to improve its performance for generating data. Second, Variational Autoencoders (VAEs) \cite{vae_text, vae, generative_lta_vae_2, generative_lta_vae_1, actor} learn to encode large quantity of data into low-dimensional latent space, and then reconstruct the original data based on that latent space representation. Finally, Diffusion Models (DMs) \cite{dm} have become popular recently due to their great performance in image synthesis tasks \cite{dalle2, imagen}. DMs work by destroying the data iteratively through Gaussian noise addition, and learning how to reverse the noise injection process by gradually denoising the sample.

This work is inspired by \cite{generative_lta_vae_2, generative_lta_vae_1} that aim to generate future human actions based on observed actions by VAEs, but refers to \cite{actor} more to condition the prediction result based on the estimated human intention. Our architecture is based on \cite{actor}, where a Transformer Encoder-Decoder based architecture \cite{attention} is designed to synthesize human motions conditioned on a categorical action through the use of distribution parameter tokens. These parameters encode the context information (past observations and intention) and serve as conditioners for the transformer decoder.

\section{Methodology}

\begin{figure}
\centering
\includegraphics[width=0.45\textwidth]{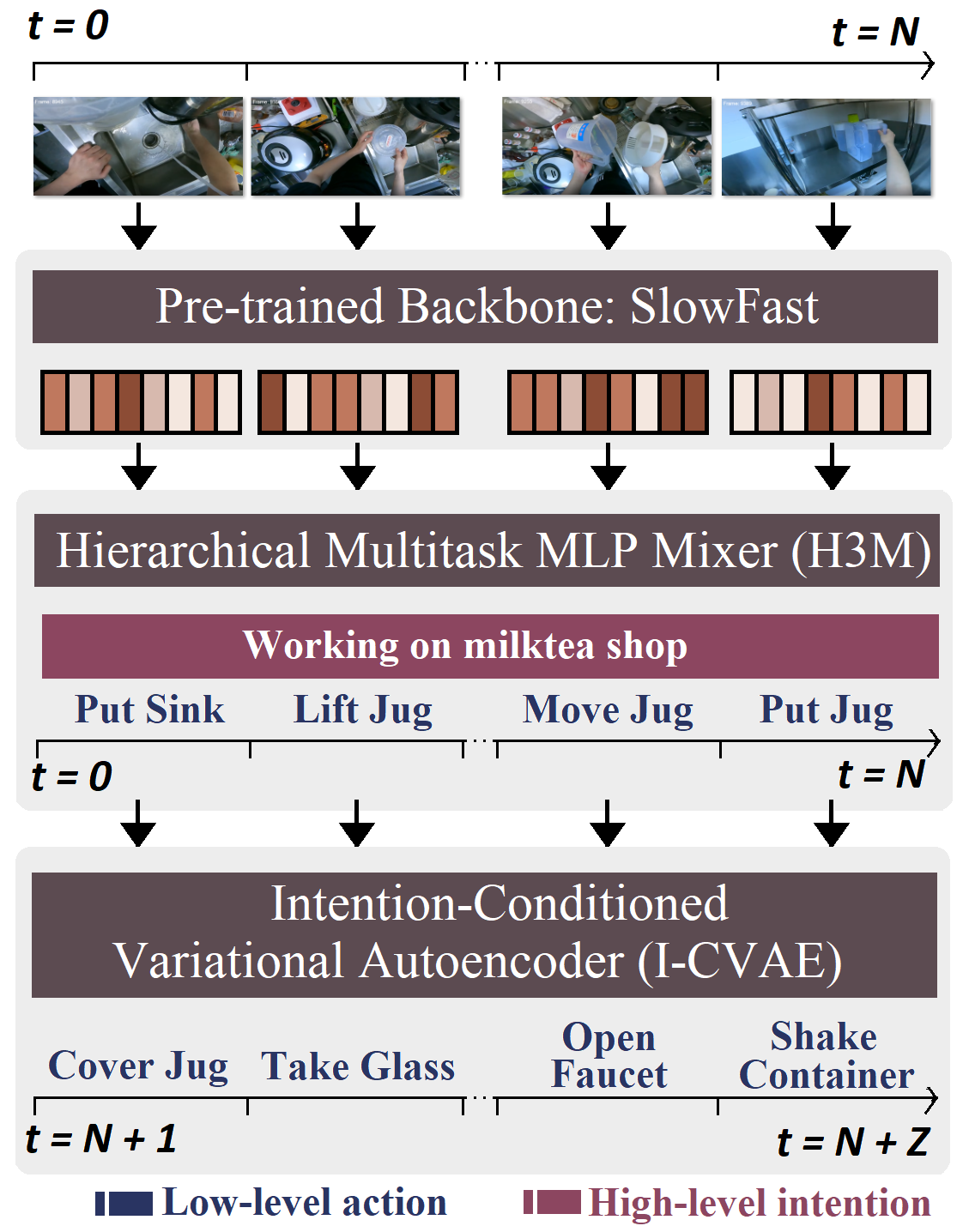}
\caption{\textbf{Overall proposed framework}. Provided pre-extracted features for $N=4$ observed videos are fed to our Hierarchical Multitask MLP Mixer model (H3M) to obtain low-level action labels and high-level intention. Results are fed into our Intention-Conditioned Variational AutoEncoder (I-CVAE) that anticipates subsequent $Z=20$ actions.}
\label{fig:overall_architecture}
\end{figure}
\par

Long-Term Anticipation of human actions needs to exploit temporal dependencies among the observed actions to generate plausible human action sequences in the future. Our two-step approach first aims at understanding the observed actions through a Hierarchical MultiTask MLP Mixer (H3M), described in Section \ref{ch:H3M} in a bottom-up approach. Then, a Transformer-based Encoder-Decoder structure \cite{attention}  is proposed for the Intention-Conditioned Variational Auto-Encoder (I-CVAE) in Section \ref{ch:ICVAE}, inspired by \cite{actor}. Fig. \ref{fig:overall_architecture} shows the overview of the proposed framework, that represents how the structure is extracted from the observed past and used to condition the future actions to be generated.

\subsection{Problem Formulation} 

Let $\mathbf{a}_t=(v_t, n_t)$ denote an action label at time $t$, which consists of verb label $v_t$ and noun label $n_t$. Then, past $N$ observed action of the camera wearer from a given untrimmed video can be represented as $\mathbf{A}_{obs} = [\mathbf{a}_1,\ldots,\mathbf{a}_N]$. Our aim is to predict the future sequence of $Z$ actions such as $\mathbf{A}_{pred}=[\mathbf{a}_{N+1},\ldots,\mathbf{a}_{N+Z}]$, by generating $K$ possible sequences $\mathbf{A}_{pred}$ to account for variations.

For a single video clip $\mathbf{V}_t=\{\mathbf{I}_{t},\ldots,\mathbf{I}_{t+T}\}$ which consists of image frames $\mathbf{I}$ from time $t$ to $t+T$, Ego4d benchmark provides a set of feature vectors $\mathbf{F} = \{\mathbf{f}_1,\ldots,\mathbf{f}_T\}$, where $\mathbf{f}_t \in \mathbb{R}^{2304}$ denotes a feature vector obtained from image frames in one-second video. Each $f$ is obtained from SlowFast architcture \cite{slowfast} pre-trained with Kinetics dataset \cite{kinetics}. We apply zero-padding in $\mathbf{F}$ to ensure the same $T$ duration for all video clips.

Then, from the observed $N$ video clips, we obtain a $N$-sequence of pre-extracted visual features $\mathbf{F}_{obs}=[\mathbf{F}_{1},\cdots, \mathbf{F}_{N}]$, which we use as an input for our H3M model. Obtained two-level outputs composes the input for I-CVAE: (i) a high-level intention prediction $\hat{I}$ is denoted as a unique label that classifies the overall goal of the camera wearer for the whole task; and (ii) a low-level action sequence prediction $\mathbf{\hat{A}}_{obs}$. Finally, both $\hat{I}$ and $\mathbf{\hat{A}}_{obs}$ are used for generating $K$ possible future action sequences $\mathbf{\hat{A}}_{pred}$.

Next, we describe our proposed two independent architectures that aim to (i) obtain the hierarchical structure of the human task and (ii) anticipate the low-level actions conditioned by the human intention.

\subsection{Hierarchical Multitask MLP Mixer (H3M)} \label{ch:H3M}

\begin{figure}
\centering
\includegraphics[width=0.43\textwidth]{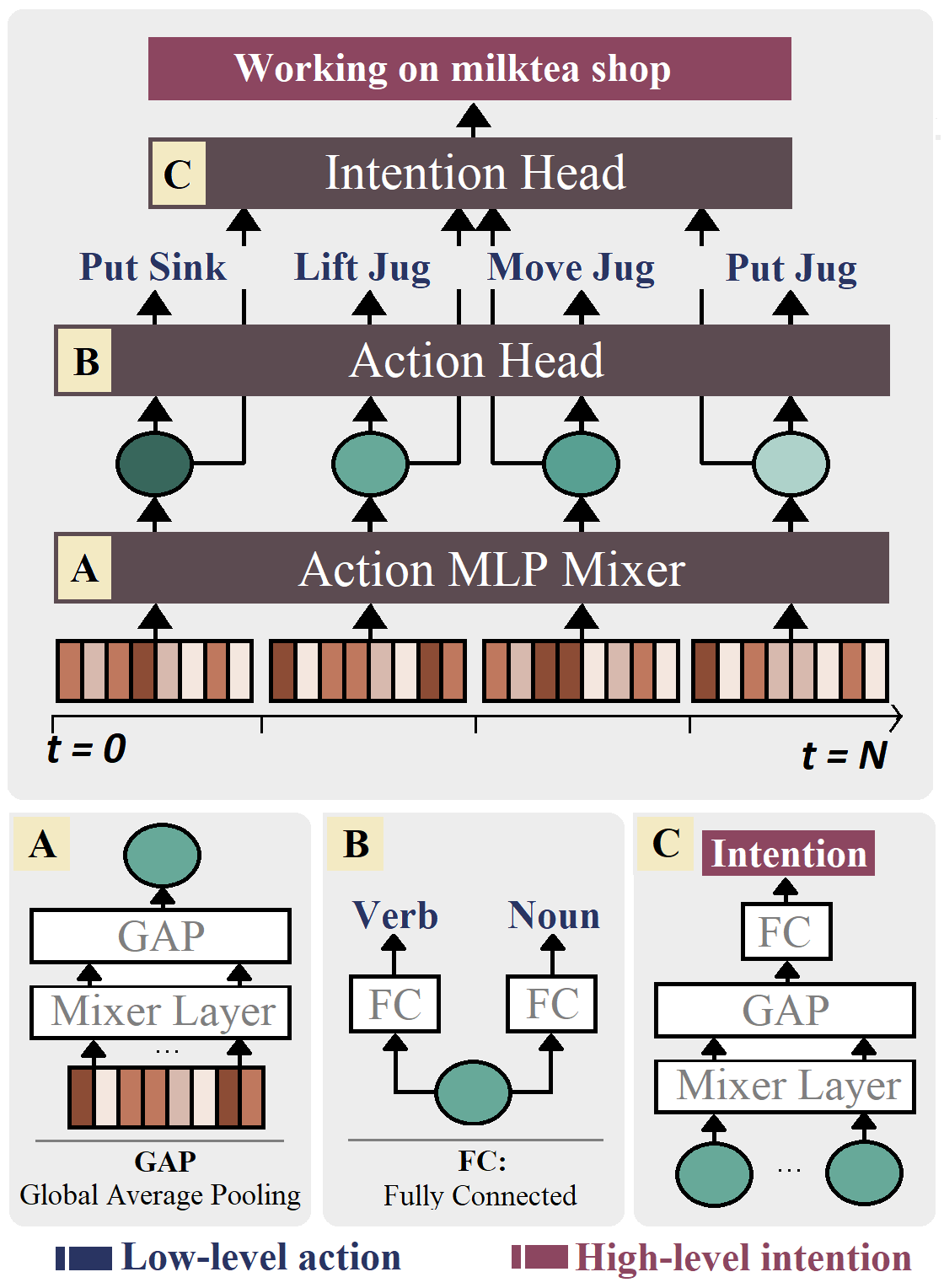}
\caption{\textbf{Detailed structure of H3M architecture.} First, pre-extracted padded features are fed into an Action MLP Mixer \cite{mlpmixer} to obtain clip level features (as green circles). These features are used (i) to obtain verb-noun pair through a fully-connected pair (action head); (ii) to obtain a video representation through the Intention MLP Mixer which is classified as an intention class. Definition of Mixer Layer is inherited from \cite{mlpmixer}.}
\label{fig:h3m}
\end{figure}

As shown in Fig. \ref{fig:h3m}, the pre-extracted features $\mathbf{F}_{obs}=[\mathbf{F}_{1},\cdots, \mathbf{F}_{N}]$ that describe the visual information of the untrimmed egocentric video are fed into a Hierarchical Multitask MLPMixer (H3M) architecture. The Action Mixer model processes $\mathbf{F}_{t}$ in parallel to encode the visual information that define a given $\mathbf{a}_t=(v_t, n_t)$. Let $\mathbf{F}_{t} \in \mathbb{R}^{T \times 2304}$, composed by $T$ feature patches $f_{t}$. Mixer layer is defined in \cite{mlpmixer} and is composed of 2 MLP blocks encapsulated between two matrix transposition operations that capture the global context of $f_{t}$. The first MLP block identifies strong temporal dependencies and mixes the data among each tokenized patched, while the second MLP block leverages spatial same-patch features. Each MLP block consists of two fully-connected layers with a GELU activation function in-between. Finally, global average pooling is applied in the $T$ dimension to project $\mathbb{R}^{T \times 2304}$ to $\mathbb{R}^{2304}$, encoding the action representation $\mathbf{x}_t$ for an observed action. The action head of the H3M projects and classifies each $\mathbf{x}_t$ as $\hat{v}_t$ and $\hat{n}_t$ through a Fully Connected layer, obtaining $N$ $\mathbf{\hat{a}}_t=(\hat{v}_t, \hat{n}_t)$ pairs to conform $\mathbf{\hat{A}}_{obs}$. Finally, the intention head applies a second MLP Mixer to leverage the global context of the sequence of action representation features $\mathbf{X}=[x_1,\cdots, x_N]$ and classifies the human intention $\hat{I}$ in the observed video. The overall hierarchical multitask classifier obtains the top-low level information from past observation. 

\par
\subsection{Intention-Conditioned Variational Autoencoder (I-CVAE)} \label{ch:ICVAE}

\begin{figure}
\centering
\includegraphics[width=0.47\textwidth]{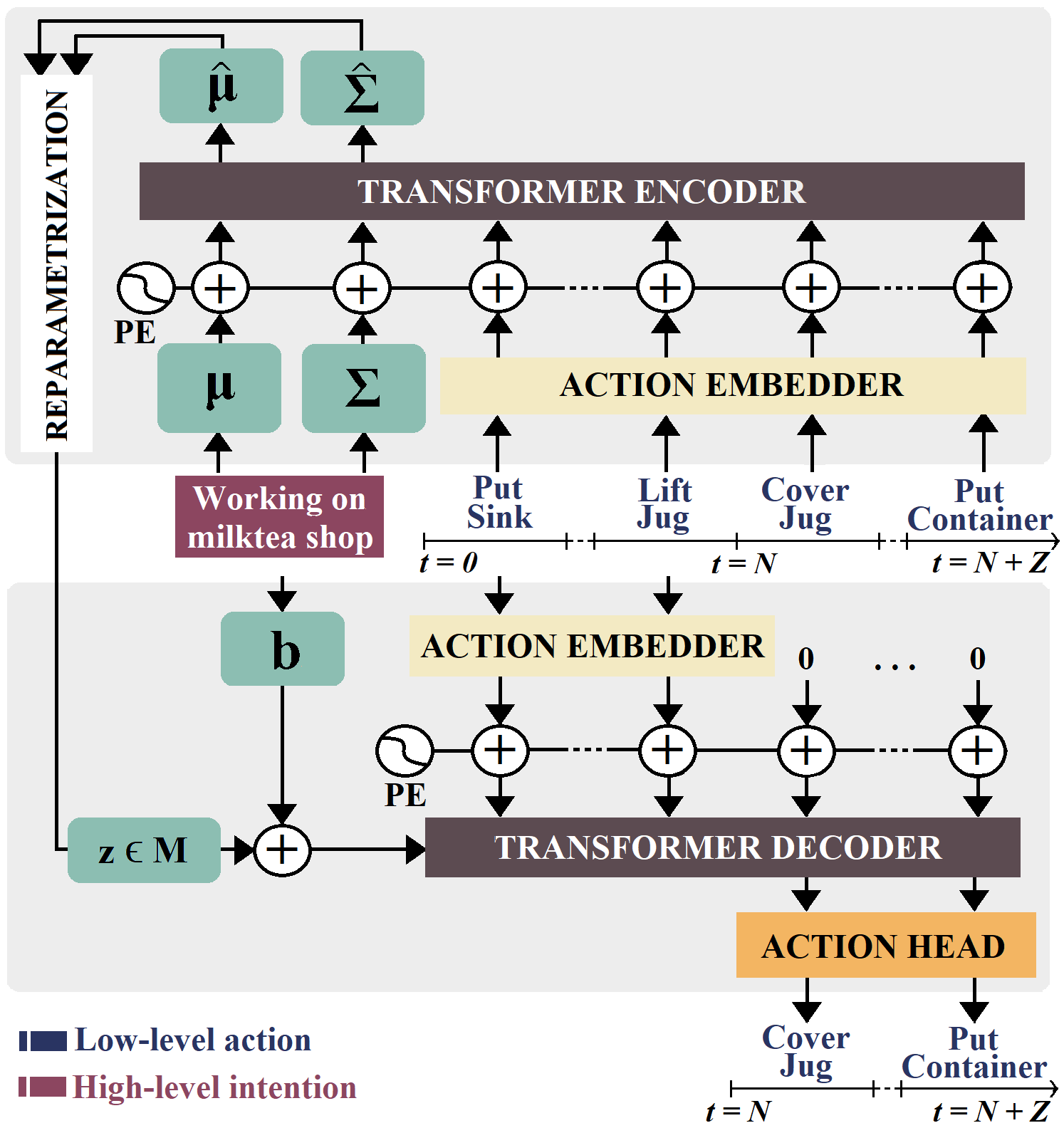}
\caption{Detailed structure of I-CVAE architecture, illustrating the encoder (top) and decoder (bottom) of our Transformer-based CVAE model. Given a sequence of $N+Z$ actions and an Intention label, the encoder outputs distribution parameters ($\hat{\mu}$ and $\hat{\Sigma}$) that encode all sequence information. Inspired by \cite{actor}, extra learnable parameters per intention are used ($\mu$ and $\Sigma$) to obtain $\hat{\mu}$ and $\hat{\Sigma}$ and sample the latent future action representation $z \in \mathbb{R}^{M}$, where $M$ is the latent dimension of the Transformer. The decoder takes a latent vector $z$, the $N$ observed actions and the intention $I$ to output the representation sequence of $Z$ actions to anticipate. $I$ is used to determine the learnable $b$. Positional Encoder (PE) gives the time-component knowledge to the decoder. Finally, an Action Head compound by two fully-connected layers projects each action representation into a verb-noun pair.}
\label{fig:icvae}
\end{figure}
\par

Fig. \ref{fig:icvae} shows the Intention-Conditioned Variational Autoencoder (I-CVAE) architecture . Encoder-Decoder Transformer structure is used to leverage temporal dependencies between past actions to anticipate future sequence of human verb-noun pairs. Due to the nature of VAE, in the training scenario the model learns the conditional probabilistic distribution in the encoder through exploiting both observed actions $\mathbf{A}_{obs}$ and future actions $\mathbf{A}_{pred}$. In its application, only the decoder block is used without any provided information of its future.

First, each action pair $\mathbf{a}_t=(v_t, n_t)$ of $\mathbf{A}_{obs}$ is projected independently through learnable embedding matrices into $\mathbf{e}_t = (e^v_t, e^n_t)$, where $e^v_t$ and $e^n_t$ consist of $d$-dimensional embedding vector for verb and noun, respectively. Therefore, the verb embeddings are represented as $\mathbf{E}^{v}_{obs} = [\mathbf{e}^{v}_{1},\ldots,\mathbf{e}^{v}_{N}]$ for the observed sequence and $\mathbf{E}^{v}_{pred} = [\mathbf{e}^{v}_{N},\ldots,\mathbf{e}^{v}_{N+Z}]$ for the future sequence (same procedure for nouns, with $\mathbf{E}^{n}_{obs}$ and $\mathbf{E}^{n}_{pred}$). To strengthen the dependencies between verb and noun as a common action, $\mathbf{E}^{v}$ and $\mathbf{E}^{n}$ are concatenated to $\mathbf{E} \in \mathbb{R}^{N \times 2d}$ in the Action Embedder. Intuitively, then, our Transformer Encoder-Decoder aims to reconstruct the $\mathbf{E}_{pred}$ in the decoder given only the $\mathbf{E}_{obs}$ and the intention label $I$ as conditions.

Inspired by \cite{actor}, a condition label $I$ is projected into extra-learnable distribution parameters $\bm{\mu}$ and $\bm{\Sigma}$ to inherit the action sequence representation after the Transformer Encoder. These tokens are prepend to the sequence of embedded representations of both the observed and forecasting actions to pool the time dimension. This whole embedded sequence are summed to a sinusoidal Positional Encoder (PE) and fed into the Transformer encoder, as can be seen in the top diagram of Fig. \ref{fig:icvae}. The obtained parameters $\hat{\mu}$ and $\hat{\Sigma}$ are then used to define a latent distribution based on the reparametrization trick from VAE. We sample $z \in \mathbb{R}^{2d}$ from this latent distribution in the decoder.

During the decoding phase, as illustrated in the bottom diagram of Fig. \ref{fig:icvae}, only the $N$ observed actions $\mathbf{A}_{obs}$ and intention label $I$ are used to condition the future generation of next $Z$ actions $\mathbf{A}_{pred}$. A learnable bias parameter $b$ is used to shift the latent representation vector $z$ to an intention-dependent space. The observed actions embeddings $\mathbf{E}_{obs}$ are appended to a zero-vector for each $Z$ actions to predict and summed with PE to form the input pattern to the Transformer Decoder. Moreover, the shift latent space is also fed in the transformer to condition the anticipation of $\mathbf{\hat{E}}_{pred}$ based on $I$. Finally, for each $t =[N+1, \ldots, N+Z]$, the predicted future actions representations $\mathbf{\hat{e}}_t$ are fed into the Action Head, which applies a fully-connected layer to classify $\mathbf{\hat{e}}_t$ into $\mathbf{\hat{a}}_t=(\hat{v}_t, \hat{n}_t)$ .

To guide I-CVAE to decode plausible actions representations, an L2 reconstruction loss is applied between $\mathbf{E}_{pred}$ and $\mathbf{\hat{E}}_{pred}$. Finally, to ensure the correct classification of the actions $\mathbf{\hat{a}}_t$, a weighted cross-entropy loss is used between $\mathbf{A}_{pred}$ and $\mathbf{\hat{A}}_{pred}$.

\section{Experiments}\label{ch:experiments}

In this chapter, we present the Ego4D Dataset \cite{ego4d} and its proposed baselines and report and compare our quantitative results. Finally, we demonstrate the effectiveness of our approach through an ablation study.

\subsection{Dataset - Ego4D}
To appropriately anticipate the future, it is necessary to understand in detail the observed actions. Human Action Recognition (HAR) from video is itself a large computer vision research field, with increasing interest over egocentric view datasets \cite{epickitchen, ego4d}. Ego4D \cite{ego4d} is the most extensive daily-life egocentric video dataset, currently available to research. It is notable that the Ego4D authors also provide pre-extracted features for each second of video. These features are obtained through SlowFast 8x8 model \cite{slowfast}. Due to the recent publication of the Ego4D dataset, only baselines results are provided for comparison. The Forecasting Benchmark from Ego4D (which includes LTA) consists of $120$ hours of annotated videos from $53$ different scenarios. The annotations provided contain $478$ noun types and $115$ verb types, with a total amount of $4756$ action classes among training and validation set. Ego4D has a long-tailed distribution both for nouns and verbs categories, resulting in a high imbalanced dataset.

\subsection{Evaluation Metrics and Baselines}
\textbf{Metrics.} Following the proposed evaluation protocols in Ego4D LTA benchmark \cite{ego4d}, we report the Edit Distance (ED) metric, as shown in Equation \ref{eq:error}, computed as the Damerau-Levenshtein distance \cite{ed1, ed2} over sequences of predictions of verbs, nouns and actions. This metric accounts for small variations in the action sequence as predicting over long time horizons is subject to uncertainty. The proposed evaluation selects the best of $K$ generated sequences according to the smallest metric. The lower the ED, the more similar the anticipated sequences to the reality. 

\vspace*{-3mm}
\begin{equation}
    \Delta E({(\hat{n}^{(j)}_{z, k}, \hat{v}^{(j)}_{z, k})}^{Z}_{z=1}, {(n^{(j)}_{z}, v^{(j)}_{z})}^{Z}_{z=1)}) 
\label{eq:error}
\end{equation}

\textbf{Baselines.} We compare our proposed framework with the baseline model proposed in Ego4D \cite{ego4d}, where a trimmed video is used to predict $K$ different plausible sequence of future actions. The baseline consists of (i) an encoder backbone for obtaining $N$ video-level features based on SlowFast \cite{slowfast}; (ii) a transformer-based aggregation module that combines the previously extracted clip-level features through self-attention mechanisms; (iii) a Multi-Head decoder network with $Z$ heads (one head per future time step) that generate sequence of future actions. $K$ future action possible sequences are generated by sampling the predicted future action distribution $K$ times. Moreover, we also compare our results with a CLIP-based model \cite{clip_ego4dlta} that leverages multi-modality to encode the visual information of the observed actions. To the best of our knowledge, no other existing work has been reported to tackle the anticipation of very long-term actions (20 actions in advance).

\subsection{Quantitative Results}\label{quantitative_results}

We report our results for the LTA task in Table \ref{tab:challenge_results} based on the test set of the Ego4D LTA dataset. In this experiment, our framework predicted the $N=6$ observed actions and the overall intention from the past, to anticipate $Z=20$ future actions by generating $K=5$ sequences. Our framework performs similar or better in all the LTA metrics defined: ED for verbs, nouns and overall actions. We claim that conditioning the generative model through the intention highly improves the ED for nouns, thus performing better in the overall action anticipation. 

\begin{table}[]
\centering
\resizebox{0.40\textwidth}{!}{%
\begin{tabular}{c|c|c|c}
\textbf{}         & \textbf{VERB}  & \textbf{NOUN}  & \textbf{ACTION}         \\ \hline
Baseline \cite{ego4d}         & \textbf{0.739} & 0.780          & 0.943                   \\ 
Video+CLIP \cite{clip_ego4dlta}          & {\ul 0.74} & {\ul 0.77}         & {\ul 0.94}\\ 

\textbf{Ours} & 0.741 &\textbf{0.739} & \textbf{0.930} \\ 
\end{tabular}%
}
\caption{Edit Distance (ED) comparison of long-term human action anticipation in Ego4D dataset. Scores are obtained directly from their reported results. Here, bold fonts denote the best result and underline denote the second-best result among all approaches. }
\label{tab:challenge_results}
\end{table}

Due to our limited computational resources, training was performed independently for each module. Next we will describe the quantitative evaluation first for the H3M module and then for I-CVAE as stand-alone models. As LTA-Ego4D Forecasting benchmark is private, the ground truth from the testing set is not provided. Therefore, to validate our hypothesis, we perform an ablation study which is based on the results obtained from validation set.

\subsubsection{\textbf{H3M}} 

Our model is able to recognize verb-noun pairs with similar performance as the baseline, as reported in Table \ref{tab:h3m_comparison}. However, our model is trained on pre-extracted features $\mathbf{F}$ for each clip, and not the image-based video clip $\mathbf{V}$ . Due to the lower dimensionality of these features ($\mathbf{F} \in \mathbb{R}^{T \times 2304}$, with $T=14$) several techniques were applied to avoid overfitting. We applied Gaussian noise injection (defined as $N$ in Table \ref{tab:h3m_comparison}) in the pre-extracted input features to improve robustness of the classifier. By adding multi-task approach and sharing inner layers among the tasks (M), we inherited implicit data augmentation: as there was a need of modelling a representation of three tasks (classifying intention, verb and noun), the model was forced to extract a better pattern from the inner shared layers to optimize results for each task. Finally, focal loss \cite{focal_loss} modulation factor ($\beta=0.99$) for the cross-entropy loss was applied to address the class imbalance (I) of intentions, verbs and nouns. The best results were obtained when an initial training was done to recognize the intention using multi-task approach, and then fine-tuning the action layers for the verb-noun recognition.

In Table \ref{tab:h3m_comparison}, it is shown that applying all techniques (M+I+N) obtains the best Top-1 accuracy results of our framework, but our model slightly decreases its performance regarding Top-5 predictions. We claim that by dealing with imbalance of the Ego4D dataset through Focal Loss, the model takes more risk by attempting to predict less-frequent actions, which impacts the accuracy. Finally, our model is limited by the low-dimensionality of inputs, which are pre-extracted visual features as mentioned above, but is still able to approximate to the baseline.

\begin{table}[]
\centering
\resizebox{0.47\textwidth}{!}{%
\begin{tabular}{c|cc|cc|cc}
 & \multicolumn{2}{c|}{\textbf{INTENTION}}       & \multicolumn{2}{c|}{\textbf{VERB}} & \multicolumn{2}{c}{\textbf{NOUN}} \\ 
\multirow{-2}{*}{\textbf{Accuracy}} &
  \multicolumn{1}{c}{\textbf{TOP1}} &
  \multicolumn{1}{c|
  }{\textbf{TOP5}} &
  \multicolumn{1}{c}{\textbf{TOP1}} &
  \multicolumn{1}{c|}{\textbf{TOP5}} &
  \multicolumn{1}{c}{\textbf{TOP1}} &
  \multicolumn{1}{c}{\textbf{TOP5}} \\ \hline
{\textbf{Ours (M+I+N)}} & \textbf{78.50}        & 93.27                 & {\ul 20.44}      & 55.05           & {\ul 19.32}      & 39.65           \\ 
{Ours (I+N)}            & {\ul 76.33}                 & {\ul 94.06}                 & 20.12            & 55.14           & 18.64            & 39.95           \\
{ Ours (N)}              & 74.92                 & \textbf{94.20}         & 20.18            & {\ul 56.02}     & 19.16            & {\ul 40.02}     \\ 
{Baseline}              & - & - & \textbf{22.06}   & \textbf{56.90}  & \textbf{20.92}   & \textbf{41.40}  \\ 
\end{tabular}
}
\caption{Performance of H3M with different training strategies, compared to the baseline using the accuracy metric. M: multitask surrogate loss (sharing weights). I: focal loss to solve class imbalance. N: noise injection. Here, bold fonts denote the best result and underlines denote the second best result among all approaches. }
\label{tab:h3m_comparison}
\end{table}

We also compare the influence of predicting intention correctly to the accuracy of action classification, illustrated in Table \ref{tab:influence}. The results show that there is a significant and direct relationship between noun and intention. By conditioning the action-level prediction framework through the intention, it is shown that the performance in terms of noun prediction is improved. As our intention label is obtained, in this work, as the `scenario' ground-truth of Ego4D, we argue that while verbs are more related to the human motion, nouns are coupled with the scenario, which refer to the environment of the human. Therefore, conditioning based on the environment has direct effect on the accuracy of noun-classification.

\begin{table}[]
\centering
\resizebox{0.35\textwidth}{!}{%
\begin{tabular}{c|cc|cc}

& \multicolumn{2}{c|}{\textbf{VERB}} & \multicolumn{2}{c}{\textbf{NOUN}} \\ 
\multicolumn{1}{c|}{\textbf{INTENTION}} &
  \multicolumn{1}{c}{\textbf{TOP1}} &
  \multicolumn{1}{c|}{\textbf{TOP5}} &
  \multicolumn{1}{c}{\textbf{TOP1}} &
  \multicolumn{1}{c}{\textbf{TOP5}} \\ \hline
\multicolumn{1}{c|}{Correct} & 20.13            & 54.32           & \textbf{19.48}   & \textbf{41.58}  \\ 
\multicolumn{1}{c|}{Error}   & \textbf{21.48}   & \textbf{57.46}  & 18.79            & 33.28           \\ 
\end{tabular}%
}
\caption{Intention influence on action classification. Scores based on the accuracy metric.}
\label{tab:influence}
\end{table}

Moreover, we further evaluate the influence of the intention as the context for our H3M model. For a given intention label, only a few verbs and nouns are observed. Then, we define an out-of-context error as predicting a verb or a noun which is unseen in a given intention. For instance, if the model predicts the action `drive bike' in a video where the human intention is `washing a dog', we claim that the model has an out-of-context error. To determine the observed classes for a given intention, we create a bag of nouns and verbs used for each. If we predict a noun that cannot be found for the current intention bag, we determine this as an out-of-context error. We can observe from our experiments that no verb is out of context, while $14.56\%$ of nouns are. These results strengthen the observation that (i) nouns and intentions have a significant relationship, (ii) verbs are less conditioned by intention and they are more related to the history of actions performed, as will be empirically shown in the next section.

\subsubsection{\textbf{I-CVAE}} 
We evaluate the performance of our standalone I-CVAE model based on the ground-truth action and intention labels provided by Ego4d dataset. We report Edit Distance (ED) for the time horizon $Z=20$ ($ED@Z=20$) as our evaluation metric for both nouns and verbs.

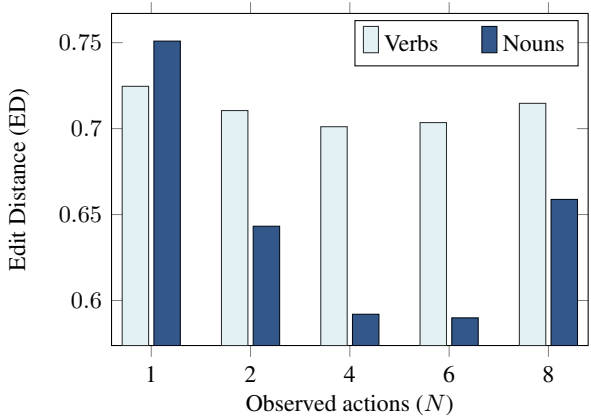
\begin{figure}[]
    \begin{tikzpicture}
        \centering

        \begin{axis}
            [ybar,
                height=6cm,
	            width=0.95\columnwidth,
                xlabel = Observed actions ($N$),
                ylabel = Edit Distance (ED),
            	style={font=\small},
                symbolic x coords={1,2,4,6,8},
                legend style={legend columns=-1},
                legend style={/tikz/every even column/.append style={column sep=0.5cm}},
                legend image code/.code={
                \draw [#1] (0cm,-0.1cm) rectangle (0.2cm,0.25cm); },
                ]
                ]
            \addplot[light_blue!20!black,fill=light_blue!80!white] coordinates {(1,0.7246) (2,0.7105) (4, 0.7011) (6,0.7035) (8,0.7147)};
            \addplot[dark_blue!20!black,fill=dark_blue!80!white] coordinates {(1,0.7509) (2,0.6432) (4, 0.5920) (6,0.5899) (8,0.6588)};
            \legend {Verbs, Nouns};
        
        \end{axis}
    \end{tikzpicture}
    \caption{Evaluation of I-CVAE trained based on different number of observed actions $N$.}
    \label{fig:observedactions}
\end{figure}

First, we study the performance of I-CVAE under the variation of the number of observed actions $N$. Fig. \ref{fig:observedactions} shows that the best results are obtained when leveraging $N=4$. Observing longer past causes the model to focus on less-relevant cues, thus collapsing in less plausible action sequences. In addition, we observe that the noun performance is inversely affected by $N$, anticipating with less confident if $N$ is lower, but not verbs. We claim that these results are caused by the different behaviours of verbs and nouns during a sequence of actions. On the one hand, verbs tend to change more frequently, but they usually appear in repeated patterns. For example, a sequence of `take-move-open-wash-close-put' is typical when `washing' an object. On the other hand, nouns are less variant as a human usually interacts with a certain object for longer duration. Reasoning that the next object that the human will interact with next depends on the environment as well as previous verbs. 

\begin{figure}[]
    \begin{subfigure}[pt]{1\columnwidth}
    \centering
    \begin{tikzpicture}
        \begin{axis}[
        	ylabel=Edit Distance (ED),
        	style={font=\small},
            title style={yshift=-1.5ex,},
        	ymin = 0.5,
        	ymax=0.80,
	    	width=0.95\columnwidth,
	    	height=5cm,
        	title=\textbf{Verbs Anticipation},
            ]

            \addplot[color=blue,mark=x] coordinates {
                (0,0.7087)
                (1,0.7208)
                (2,0.7257)
                (3,0.73)
                (4,0.7342)
                (5,0.7382)
                (6,0.7424)
                (7,0.7465)
                (8,0.7502)
                (9,0.7533)
                (10,0.756)
                (11,0.7582)
                (12,0.7596)
                (13,0.7613)
                (14,0.7629)
                (15,0.7643)
                (16,0.7657)
                (17,0.7671)
                (18,0.7683)
                (19,0.7694)
                };
            
            \addplot[color=orange,mark=*] coordinates {
                (0,0.5436)
                (1,0.57)
                (2,0.585)
                (3,0.5985)
                (4,0.6088)
                (5,0.6178)
                (6,0.6263)
                (7,0.6337)
                (8,0.6404)
                (9,0.6464)
                (10,0.652)
                (11,0.6573)
                (12,0.6628)
                (13,0.6677)
                (14,0.6721)
                (15,0.6764)
                (16,0.6802)
                (17,0.6837)
                (18,0.6867)
                (19,0.6895)
                };
            
            \addplot[color=teal,mark=square*] coordinates {
                (0,0.5103)
                (1,0.5349)
                (2,0.5442)
                (3,0.5538)
                (4,0.5615)
                (5,0.5692)
                (6,0.5767)
                (7,0.5834)
                (8,0.5894)
                (9,0.5947)
                (10,0.6005)
                (11,0.6064)
                (12,0.6118)
                (13,0.6172)
                (14,0.6224)
                (15,0.6274)
                (16,0.6323)
                (17,0.6366)
                (18,0.6403)
                (19,0.6436)
                };
            
            \addplot[color=red,mark=otimes*] coordinates {
                (0,0.5145)
                (1,0.5357)
                (2,0.5457)
                (3,0.5546)
                (4,0.5631)
                (5,0.5706)
                (6,0.5771)
                (7,0.5828)
                (8,0.5884)
                (9,0.5936)
                (10,0.5989)
                (11,0.6041)
                (12,0.609)
                (13,0.614)
                (14,0.6189)
                (15,0.6229)
                (16,0.6265)
                (17,0.6299)
                (18,0.6327)
                (19,0.6352)
                };
            
            \addplot[color=black,mark=triangle*] coordinates {
                (0,0.5777)
                (1,0.5922)
                (2,0.5989)
                (3,0.606)
                (4,0.611)
                (5,0.6157)
                (6,0.621)
                (7,0.6261)
                (8,0.6305)
                (9,0.6342)
                (10,0.6379)
                (11,0.6412)
                (12,0.644)
                (13,0.6468)
                (14,0.6488)
                (15,0.6511)
                (16,0.6533)
                (17,0.6551)
                (18,0.6568)
                (19,0.6584)
                };
            
        \end{axis}
    \end{tikzpicture}
    \label{fig:sub1}
\end{subfigure}
    
\begin{subfigure}[pt]{1\columnwidth}
    \centering
    \begin{tikzpicture}
    \begin{axis}[
    	xlabel=Future action step ($t$),
    	ylabel=Edit Distance (ED),
    	ymin = 0.5,
    	ymax=0.80,
    	style={font=\small},
        title style={yshift=-1.5ex,},
    	width=0.95\columnwidth,
    	height=5cm,
        legend pos= south east,
    	title=\textbf{Nouns Anticipation}
        ]
                \addplot[color=blue,mark=x] coordinates {
                (0,0.62)
                (1,0.6734)
                (2,0.6952)
                (3,0.7059)
                (4,0.7135)
                (5,0.7186)
                (6,0.7225)
                (7,0.7254)
                (8,0.7285)
                (9,0.7314)
                (10,0.7333)
                (11,0.7356)
                (12,0.7377)
                (13,0.7393)
                (14,0.7408)
                (15,0.7424)
                (16,0.7436)
                (17,0.7444)
                (18,0.7457)
                (19,0.7465)
                };
                
                \addplot[color=orange,mark=*] coordinates {
                (0,0.613)
                (1,0.661)
                (2,0.6826)
                (3,0.6938)
                (4,0.7002)
                (5,0.7052)
                (6,0.7089)
                (7,0.7127)
                (8,0.716)
                (9,0.7186)
                (10,0.7203)
                (11,0.7228)
                (12,0.7248)
                (13,0.7268)
                (14,0.7286)
                (15,0.7301)
                (16,0.7317)
                (17,0.7332)
                (18,0.7343)
                (19,0.7357)
                };
                
                \addplot[color=teal,mark=square*] coordinates {
                (0,0.5905)
                (1,0.6451)
                (2,0.6653)
                (3,0.6757)
                (4,0.6847)
                (5,0.6903)
                (6,0.6956)
                (7,0.7013)
                (8,0.7044)
                (9,0.7079)
                (10,0.7107)
                (11,0.713)
                (12,0.7158)
                (13,0.7177)
                (14,0.7196)
                (15,0.7217)
                (16,0.7235)
                (17,0.7248)
                (18,0.7266)
                (19,0.7281)
                };
                
                \addplot[color=red,mark=otimes*] coordinates {
                (0,0.6003)
                (1,0.651)
                (2,0.6719)
                (3,0.6825)
                (4,0.689)
                (5,0.6939)
                (6,0.6982)
                (7,0.7024)
                (8,0.7061)
                (9,0.7087)
                (10,0.7115)
                (11,0.7138)
                (12,0.7163)
                (13,0.7181)
                (14,0.7201)
                (15,0.7218)
                (16,0.7235)
                (17,0.725)
                (18,0.7264)
                (19,0.7275)
                };
                
                \addplot[color=black,mark=triangle*] coordinates {
                (0,0.6119)
                (1,0.6521)
                (2,0.6712)
                (3,0.6856)
                (4,0.693)
                (5,0.6994)
                (6,0.7026)
                (7,0.7067)
                (8,0.71)
                (9,0.7134)
                (10,0.7155)
                (11,0.7178)
                (12,0.7198)
                (13,0.7218)
                (14,0.7231)
                (15,0.7245)
                (16,0.7254)
                (17,0.7265)
                (18,0.7275)
                (19,0.7283)
                };

            \legend{N=1, N=2, N=4, N=6, N=8}
        \end{axis}
        \end{tikzpicture}
        \label{fig:sub2}
\end{subfigure}
\caption{Edit Distance (ED) at each time step $t$ depending on the number of observed actions $N$.}
\label{fig:variationintime}

\end{figure}
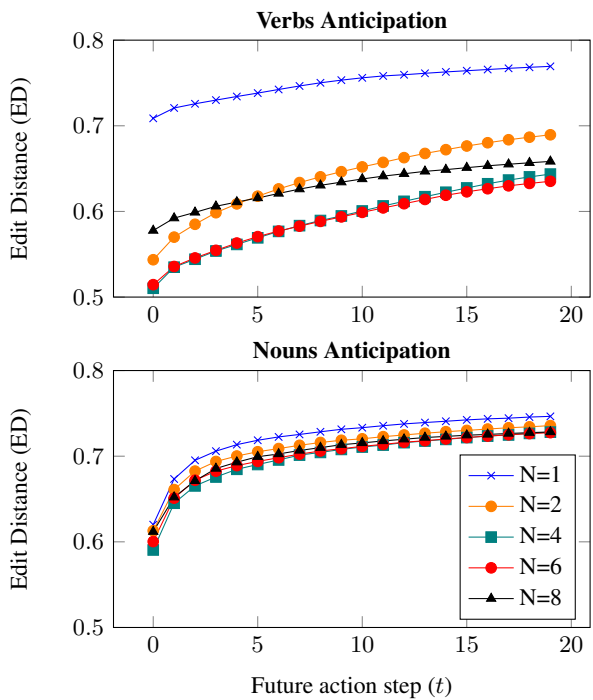

This phenomenon is also shown in Fig. \ref{fig:variationintime}, where the $ED$ for different $N$ is represented at each of the $Z=20$ time-horizons to forecast. Observing longer sequences ($N=8$) increases the variance of observed actions, which leads the model to worsen the short-term anticipation. This stresses the need of exploiting temporal dependencies using our Transformer-based model. However, only accounting the last action performed ($N=1$) caused implausible anticipations, mostly of nouns, due to the lack of context. Moreover, Fig. \ref{fig:variationintime} also illustrates how the longer time horizon estimation affects each action component differently: nouns performance is reduced linearly with each timestep compared to the logarithmic reduction of verbs performance. To sum up, we have shown the two long-standing challenges in human long-term action anticipation: (i) learning the correct pattern of human behaviours, which is highly influenced by understanding the repeating patterns of verbs in a task; and (ii) exploring future tasks to perform in a given environment, based on leveraging visual cues to provide subsequent nouns that a human is going to interact with.

\subsubsection{\textbf{H3M + I-CVAE}} 
Finally, we investigate the performance of our whole framework based on the end-to-end evaluation. First, H3M classifies the actions and the intention from the observed clips. Then, based on these predictions, our I-CVAE model anticipates the $Z=20$ actions in the future. In Table \ref{tab:end-to-end} we evaluate the LTA task in the test set for different $N$ and under the influence of the intention as a condition. Results strengthen the importance of using the intention as a guideline to make more realistic anticipations. Therefore, we are able to confirm our hypothesis of using the intention as a high-level task knowledge to narrow the arbitrariness of the future. These results are also aligned with the different behaviours of verbs and nouns previously discussed: using $N=4$ and $N=6$ provides sufficient context for the model to understand the task at hand. Observing higher number of actions causes the model to overfit.

\begin{table}[]
\centering
\resizebox{0.45\textwidth}{!}{%
\begin{tabular}{c|cc|cc|cc}
\multicolumn{1}{c|}{} & \multicolumn{2}{c|}{\textbf{ED@20 verbs}} & \multicolumn{2}{c|}{\textbf{ED@20 nouns}} & \multicolumn{2}{c}{\textbf{ED@20 Action}} \\ 
\textbf{N} & \textbf{I \cmark} & \textbf{I \xmark} & \textbf{I \cmark} & \textbf{I 
\xmark} & \textbf{I \cmark} & \textbf{I \xmark} \\ \hline
2 & 0.743 & 0.762 & 0.747 & 0.786 & 0.932 & 0.953 \\
4 & 0.742 & 0.751 & \textbf{0.736} & 0.765 & 0.932 & 0.943 \\
6 & \textbf{0.741} & 0.748 & 0.740 & 0.753 & \textbf{0.930} & 0.938 \\
8 & 0.747 & 0.755 & 0.747 & 0.758 & 0.935 & 0.939
\end{tabular}%
}
\caption{Comparison of ED@20 for verbs, nouns and actions in the test set depending on the use of the intention \mbox{(I \cmark)} or not (I \xmark) in the end-to-end approach: I-CVAE uses the H3M predictions from the $N$ observed clips. } 
\label{tab:end-to-end}
\end{table}



\section{Conclusion}

In this work, we propose a two-module framework, which consists of a Hierarchical Multitask MLP Mixer (H3M) and Intention-Conditioned Variational Autoencoder (I-CVAE), to efficiently exploit human intention as a condition to increase the confidence of the model for long-term human action sequence anticipation. Our H3M module leverages hierarchy structure to classify the observed human actions and intentions. Then, by conditioning the future through our I-CVAE, our framework anticipates better the long-term action sequence. These results reinforce the importance of defining different level of abstractions for task contextualization. Experiments demonstrate that conditioning the model through the human intention has a direct effect when improving realistic anticipation. Finally, we conduct an extensive ablation study to investigate the differences between anticipating nouns and verbs. We describe these behaviors with different pattern observed in the Ego4D dataset, which are essential to tackle the Long-Term Anticipation (LTA). Then, our work outperforms the state-of-the-art in the Ego4D LTA task, mainly when forecasting nouns and actions. Our work ranked first in both CVPR@2022 and ECCV@2022 Ego4D LTA Challenge.

{\small
\bibliographystyle{ieee_fullname}
\bibliography{egbib}
}

\end{document}